\documentclass[letterpaper]{article}
\usepackage{aaai2026}
\usepackage{times}
\usepackage{helvet}
\usepackage{courier}
\usepackage{tcolorbox}
\usepackage[hyphens]{url}
\usepackage{graphicx}
\usepackage{natbib}
\usepackage{caption}
\usepackage{booktabs}
\usepackage{amsmath}
\usepackage{xcolor}

\title{The Triangle of Similarity: A Multi-Faceted Framework for\\Comparing Neural Network Representations}

\author{
    Olha Sirikova\textsuperscript{\rm 1}\thanks{Equal contribution.},
    Alvin Chan\textsuperscript{\rm 2}\footnotemark[1]
}
\affiliations{
    \textsuperscript{\rm 1}Taras Shevchenko National University of Kyiv\\
    \textsuperscript{\rm 2}Nanyang Technological University\\
    olhasiri@gmail.com, guoweialvin.chan@ntu.edu.sg
}

\begin{document}
\maketitle

\begin{abstract}
Comparing neural network representations is essential for understanding and validating models in scientific applications. Existing methods, however, often provide a limited view. We propose the \textit{Triangle of Similarity}, a framework that combines three complementary perspectives: static representational similarity (CKA/Procrustes), functional similarity (Linear Mode Connectivity or Predictive Similarity), and sparsity similarity (robustness under pruning). Analyzing a range of CNNs, Vision Transformers, and Vision-Language Models using both in-distribution (ImageNetV2) and out-of-distribution (CIFAR-10) testbeds, our initial findings suggest that: (1) architectural family is a primary determinant of representational similarity, forming distinct clusters; (2) CKA self-similarity and task accuracy are strongly correlated during pruning, though accuracy often degrades more sharply; and (3) for some model pairs, pruning appears to regularize representations, exposing a shared computational core. This framework offers a more holistic approach for assessing whether models have converged on similar internal mechanisms, providing a useful tool for model selection and analysis in scientific research.
\end{abstract}

\section{Introduction}

As deep neural networks become integral to scientific discovery, with applications ranging from medical imaging \citep{litjens2017deep} to particle physics \citep{radovic2018machine}, a core question of interpretability remains: \textit{When do different models learn similar underlying concepts?} By tackling this question, we can gain deeper insights into the fundamental nature of these models and the data they are trained on, while leveraging the  expressivity of black-box models.

An increasing body of research aims to compare the representations of deep models from different perspectives. For instance, Centered Kernel Alignment (CKA) \citep{kornblith2019similarity} provides a straightforward way of comparing the geometry of representation spaces, while Linear Mode Connectivity (LMC) \citep{draxler2018essentially} probes whether two models exist in the same functional basin of the loss landscape. Critically, both analyze models as static entities, overlooking how their relationship changes under stress, such as the systematic removal of parameters via pruning. These methods offer different views of the underlying model similarity and can be grouped into three categories: \textbf{(i)} \textit{static view} (CKA), \textbf{(ii)} \textit{functional view} (LMC), and \textbf{(iii)} \textit{sparsity view} (parameter pruning). Conceptually, the static view tries to capture \textit{what the representations look like}, the functional perspective offers insights about the \textit{similarity between models in the weight space} and the sparsity view informs us about the \textit{robustness of the similarity}.

\begin{figure}[t]
\centering
\includegraphics[width=0.8\columnwidth]{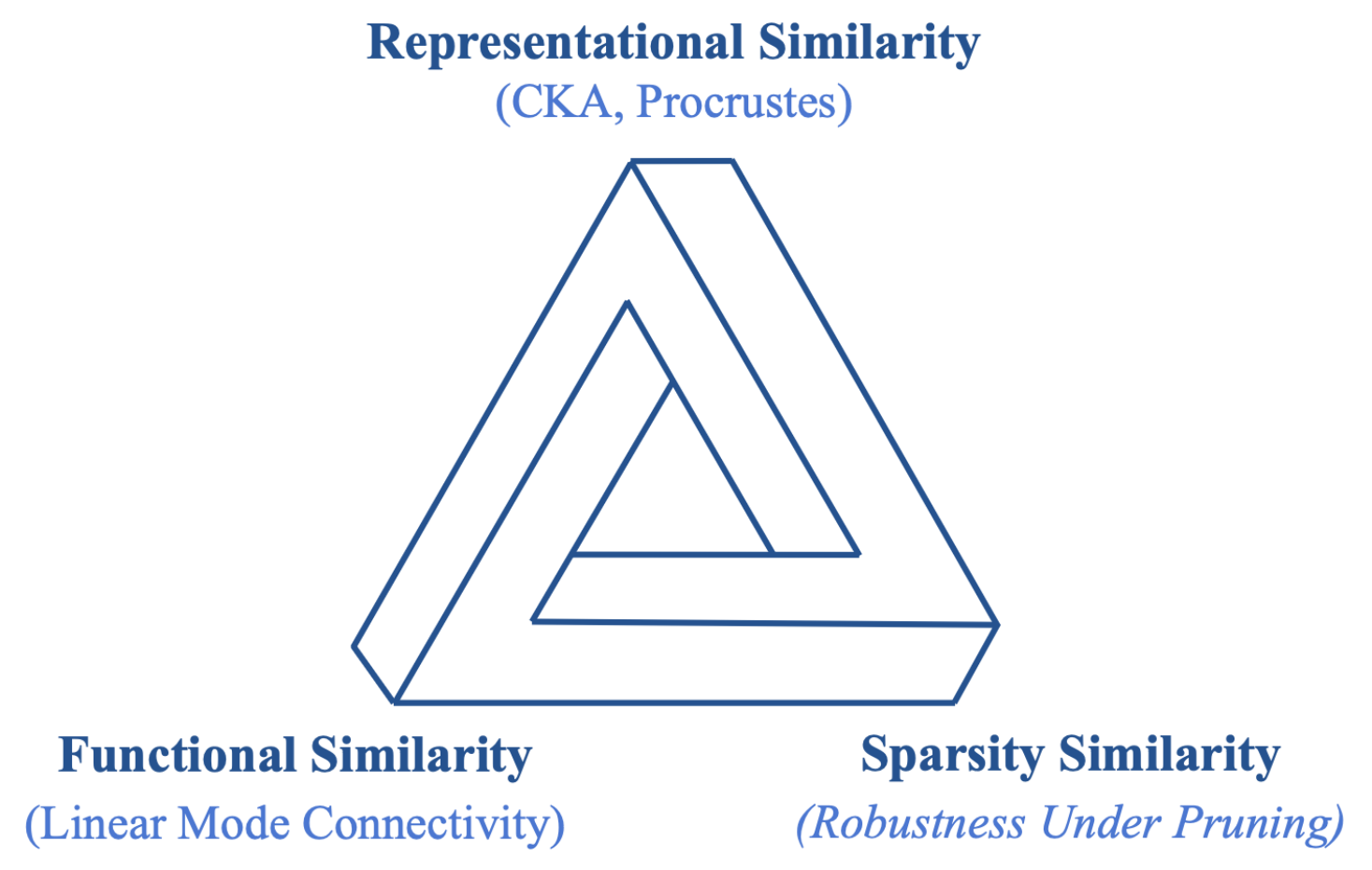} 
\caption{\textbf{The Triangle of Similarity Framework.} Our approach integrates three complementary views to form a holistic comparison: (1) static similarity of internal representations, (2) functional similarity in weight or output space, and (3) sparsity similarity as models undergo stress via pruning.}
\label{fig:conceptual_triangle}
\end{figure}

While these methods are useful on their own, a holistic understanding of model similarity requires synthesizing these perspectives. To this end, we introduce the \textit{Triangle of Similarity} (shown in Figure \ref{fig:conceptual_triangle}), a framework that aims to integrate these three analytical pillars to create a more comprehensive fingerprint of model relationships. Through this framework, we not only investigate the landscape of modern vision models but also provide a quantitative analysis of how these views interrelate. Our work aims to provide a practical toolkit and initial baselines for researchers to better understand the relationships between models, validate findings across different datasets, and make more principled decisions about model selection and transfer learning.

\section{Related Work}

\subsection{Representational Similarity Analysis}
Comparing neural representations has evolved from early methods like Canonical Correlation Analysis (CCA) \citep{raghu2017svcca, morcos2018insights} to more robust metrics. CKA \citep{kornblith2019similarity} has become a standard due to its invariance to orthogonal transformations. Orthogonal Procrustes analysis \citep{schonemann1966generalized} offers a complementary view of geometric alignment \citep{gower1975generalized}. Our work does not propose a new metric but instead integrates these established static measures into a broader, multi-faceted analytical framework, a topic recently surveyed in \citep{klabunde2025similarity}.

\subsection{Loss Landscape Geometry}
The finding that independently trained networks of the same architecture can often be connected by a high-accuracy path in weight space \citep{draxler2018essentially} has spurred significant research into the geometry of loss landscapes \citep{mirzadeh2020linear}. This area primarily focuses on model merging and ensembling. We adapt LMC as a diagnostic tool to measure the functional barrier between two models, contextualizing it with representational and behavioral analyses.

\subsection{Network Pruning}
The Lottery Ticket Hypothesis \citep{frankle2018the} showed that dense networks contain sparse, trainable subnetworks. This has led to highly efficient pruning methods \citep{frantar2023sparse}.  Research in this area typically focuses on model compression and efficiency. We utilize pruning not for compression, but as a comparative analytical tool. By observing how the relationship between two models changes as they are sparsified, we can infer the nature of their shared representations, which touches on the concept of robust, Platonic representations \citep{huh2024platonic}.

\section{Methodology}

\subsection{The Triangle of Similarity Framework}

We define our framework across any deep learning model architecture that consists of one or more hidden layers. Thus, for any pair of models ($M_A$, $M_B$), we construct a three-panel analysis that provides a holistic perspective on model similarity: (i) \textbf{static representational view}, (ii) \textbf{functional view} and (iii) \textbf{sparsity view}.

\paragraph{Panel 1: Static Representational View}
This view aims to capture the geometric structure of learned features at rest. Practically, we consider the static view on model similarity by extracting layer-wise activations and compute a full similarity matrix using CKA \citep{kornblith2019similarity} for relational geometry and Procrustes \citep{schonemann1966generalized} for geometric alignment. This establishes the baseline similarity across the models' hierarchies.

\paragraph{Panel 2: Functional View}
In contrast with the representational view, this perspective assesses whether the models compute the same function, either in the weight space or output predictions. The functional view analysis we perform as part of our framework adapts based on the architectural differences between $M_A$ and $M_B$. In the scenario where the two models \textbf{have the same architecture}, we evaluate the accuracy along the linear interpolation path $\theta(\alpha) = (1-\alpha)\theta_A + \alpha\theta_B$ , introduced in LMC \citep{draxler2018essentially}, where $\theta_A$ and $\theta_B$ represent the parameters of $M_A$ and $M_B$ respectively. Intuitively, a low-error path suggests that the models found functionally equivalent solutions. If the two models \textbf{have different architectures}, we instead compute the Predictive Similarity. This similarity is computed by obtaining softmax predictions from both models on a test set and calculating the Jensen-Shannon Divergence (JSD) between their average prediction distributions. A score of $\text{JSD}$ close to $0$ indicates the models make similar predictions.

\paragraph{Panel 3: Sparsity View}
By subjecting models to "stress" via pruning, this view reveals the robustness and shared core of their representations. In practice, we progressively prune both models using global magnitude pruning at various sparsity levels $s$. We then track two key trends: (1) \textbf{individual model accuracy} $\text{acc}(M_A^{(s)})$, $\text{acc}(M_B^{(s)})$ and (2) \textbf{cross-model similarity over layers} $\text{sim}(M_A^{(s)}, M_B^{(s)})$.

These metrics across different sparsity levels indicate whether models share a robust, sparse computational core or if their similarity is an artifact of over-parameterization.

\section{Results}

\subsection{Experimental Protocol}

\paragraph{Models} In order to test our Triangle of Similarity we analyze models from three architectural families:
\begin{itemize}
    \item \textbf{CNNs:} ResNet18 (RN18) and ResNet50 (RN50) \citep{he2016resnet}, including variants trained on ImageNet, CIFAR-10, and with random initialization.
    \item \textbf{Vision Transformers (ViTs):} ViT-Tiny-Patch16 (ViT-TP16) \citep{dosovitskiy2020vit} and DeiT-Tiny-Patch16 (DeiT-TP16) \citep{touvron2021deit}.
    \item \textbf{Vision-Language Models (VLMs):} DINOv2-Base \citep{oquab2023dinov2}, CLIP-ViT-B/32 (CLIPP32) \citep{radford2021clip}, BLIP-ViT-B/16 (BLIP16) \citep{li2022blip}, and LLaVA-1.5-7B (LLaVA, vision tower only) \citep{liu2023llava}.
\end{itemize}

\noindent By considering a wide range of models with different inductive biases and training dynamics, we provide an initial baseline for understanding the relationship between model families as well as insights useful for practitioners. 

\paragraph{Data \& Evaluation} We employ a two-tier evaluation strategy to ensure robust comparisons:
\begin{itemize}
    \item \textbf{Tier 1 (Static Similarity, Figure 2):} We use 5,000 images from the CIFAR-10 test set (upscaled to model resolution). Representational similarity metrics like CKA measure geometric alignment, which is relatively invariant to input distribution \citep{kornblith2019similarity}. Using CIFAR-10 serves dual purposes: (i) testing the robustness of architectural clustering to out-of-distribution data, and (ii) enabling rapid prototyping.
    
    \item \textbf{Tier 2 (Dynamic Analysis, Figures 3–5):} We use the ImageNetV2 subset (2,000 images, classes 0–200) for pruning and cross-view analyses where task-specific accuracy is measured. This ensures performance metrics reflect the models' intended evaluation domain.
\end{itemize}

\subsection{The Architectural Divide in Similarity}

\begin{figure}[h!]
\centering
\includegraphics[width=0.8\columnwidth]{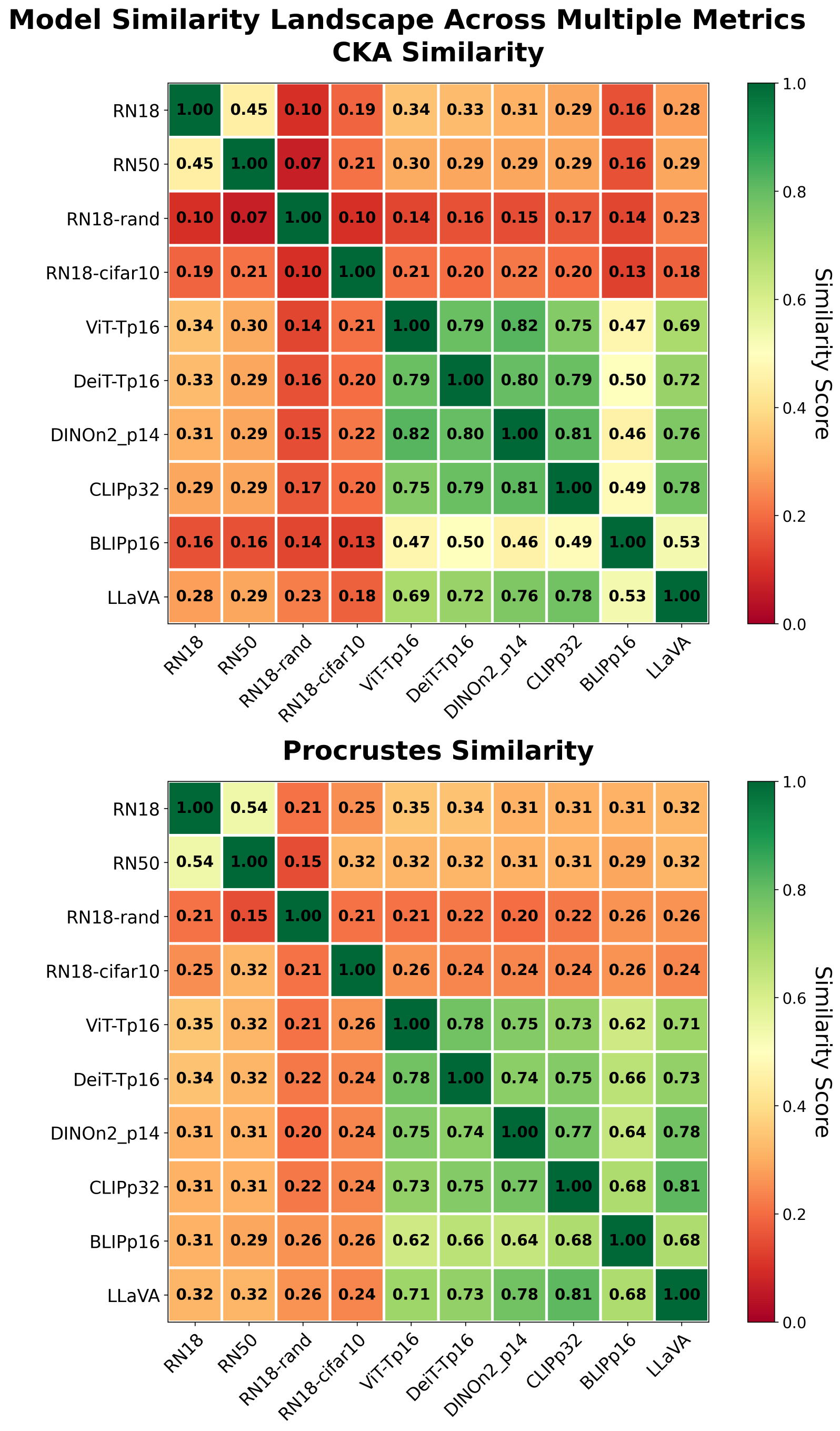} 
\caption{\textbf{Global similarity landscape on OOD data.} CKA (top) and Procrustes (bottom) heatmaps evaluated on CIFAR-10 images reveal that architectural clustering remains robust even when processing out-of-distribution data. High similarity is observed within CNNs and within Transformers, while cross-family similarity remains low.}
\label{fig:landscape}
\end{figure}

Our first analysis maps the overall similarity between models. Figure \ref{fig:landscape} shows the pairwise CKA and Procrustes similarity scores, revealing clear structural patterns.

\begin{tcolorbox}[colback=blue!2!white,leftrule=2.5mm,size=title]
\textbf{Finding 1.}  \textit{Architectural family is the primary predictor of representational similarity.}
\label{question1}
\end{tcolorbox}
The heatmaps show distinct blocks of high similarity. The Transformer-based models (ViT, DeiT, DINO, CLIP, BLIP, LLaVA) form a highly coherent cluster, indicating a shared representational format dictated by the self-attention mechanism. The CNNs form a separate, less tight cluster. Similarity between architectures from different families (e.g., RN50 vs. DINOV2) is consistently lower than within-family similarity (e.g., DeiT-TP16 vs. ViT-TP16). This provides a quantitative baseline for the known architectural differences.

\subsection{Pruning's Effect on Accuracy and Self-Similarity}

\begin{figure}[h!]
\centering
\includegraphics[width=0.85\columnwidth]{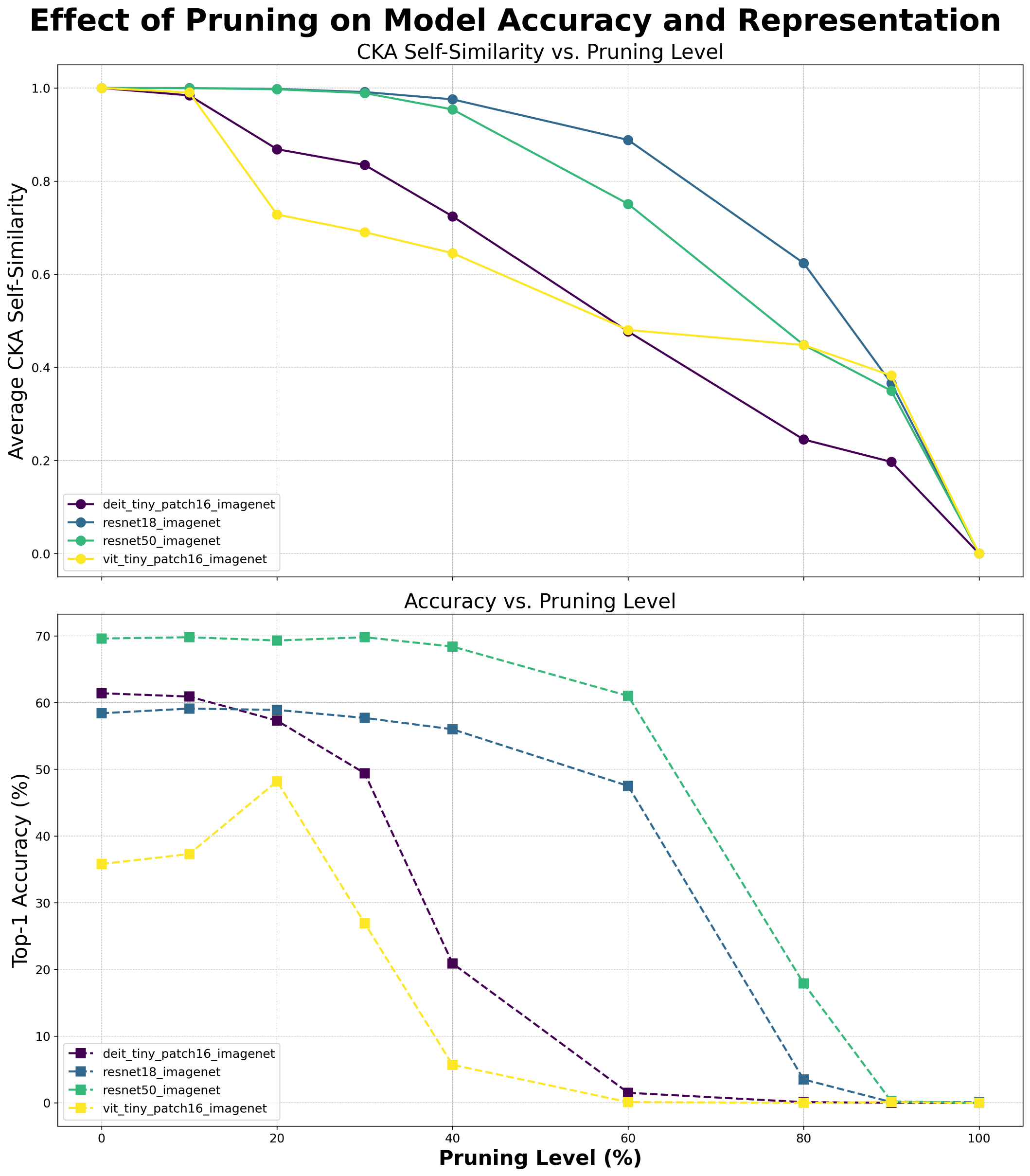} 
\caption{\textbf{Effect of Pruning.} \textbf{Top:} CKA self-similarity (pruned vs. original model) degrades as pruning increases. \textbf{Bottom:} Top-1 accuracy degrades more sharply than self-similarity, especially for Transformers like DeiT and ViT.}
\label{fig:pruning_effect}
\end{figure}

To understand the dynamic behavior of models, we analyzed their performance and representational integrity under pruning. Figure \ref{fig:pruning_effect} plots CKA self-similarity (a model against its pruned self) and Top-1 accuracy versus the pruning level.

\begin{tcolorbox}[colback=blue!2!white,leftrule=2.5mm,size=title]
\textbf{Finding 2.}  \textit{Task accuracy is more sensitive to pruning than representational structure.}
\label{question1}
\end{tcolorbox}

 For all models, CKA self-similarity declines, often maintaining a score above 0.8 even at 40\% sparsity. In contrast, accuracy (dashed lines) often experiences a much steeper drop. This is particularly pronounced for the Vision Transformers (DeiT, ViT), where accuracy falls sharply while the CKA curve remains relatively high. This suggests that while many weights are crucial for achieving the final percentage points of accuracy, the core representational structure of the model is more resilient and distributed across a larger set of parameters. The close coupling of the CKA and accuracy curves confirms that as a model's representations diverge from their original state, performance predictably degrades.

\subsection{Divergence of Functional and Representational Stability}

To dissect the model's response to stress, we compare the functional distance against the representational distance  as the model is progressively pruned. The comparison contrasts how weight space functionality (Panel 2 concept, adapted to $M$ vs. $M^{(s)}$) and internal structure (Panel 1 concept, adapted to $M$ vs. $M^{(s)}$) react to the Sparsity View stress.

\begin{figure}[h!]
\centering
\includegraphics[width=0.85\columnwidth]{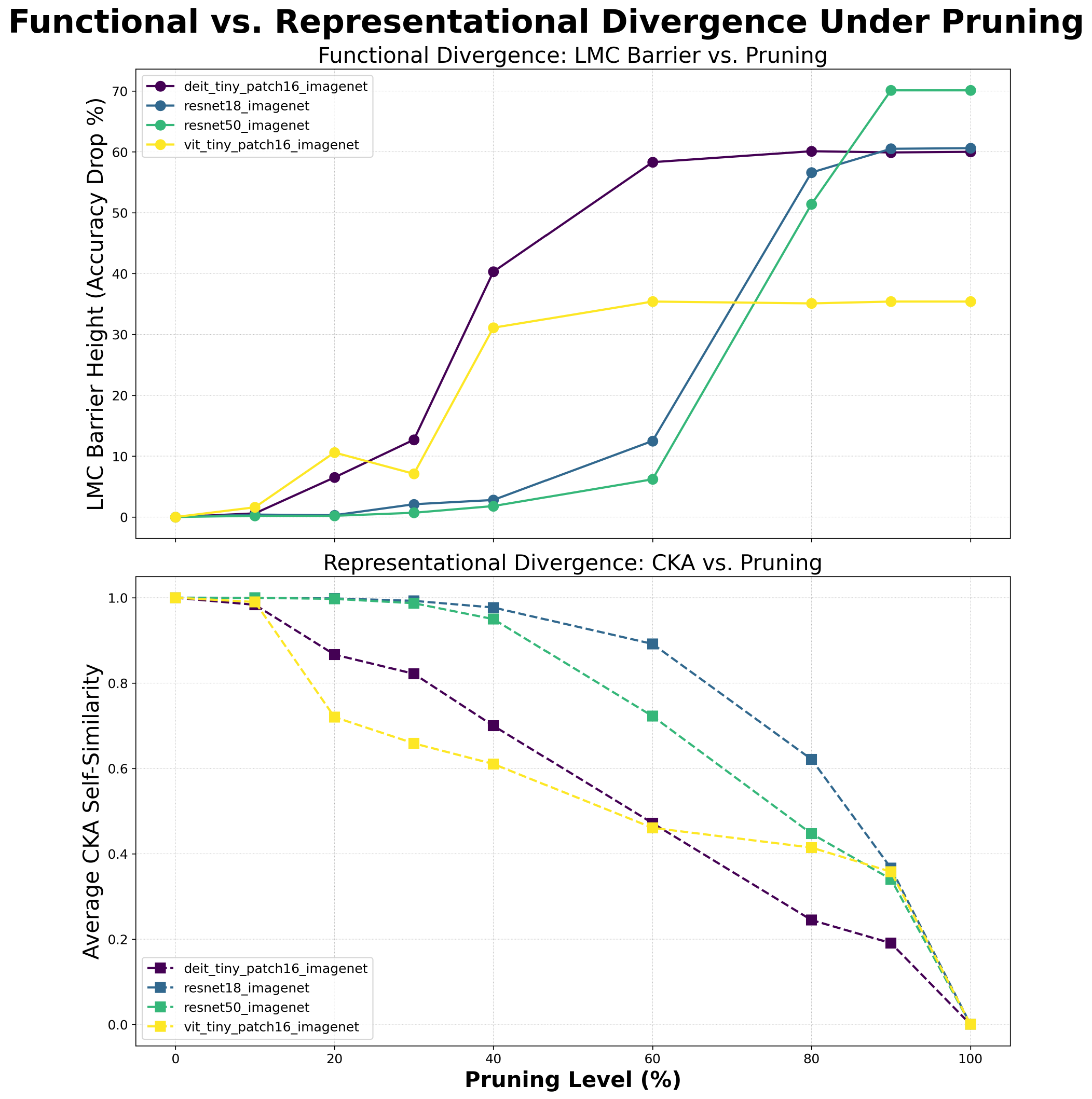} 
\caption{\textbf{Functional vs. Representational Divergence Under Pruning.} The top panel shows the Linear Mode Connectivity (LMC) Barrier Height (Functional Divergence) between the original and pruned model ($M$ vs. $M^{(s)}$). The bottom panel shows the Average CKA Self-Similarity (Representational Divergence). The LMC Barrier rises more sharply and at lower sparsity levels than the CKA Self-Similarity drops, suggesting functional proximity is a more fragile property than representational structure.}
\label{fig:lmc_cka_pruning}
\end{figure}

Figure \ref{fig:lmc_cka_pruning} presents this combined view. The top panel shows the LMC Barrier, which quantifies the minimum accuracy drop encountered along the linear path connecting the unpruned model $M$ to its pruned variant $M^{(s)}$. A zero barrier means the two models are functionally equivalent and lie in the same local minimum.

\begin{tcolorbox}[colback=blue!2!white,leftrule=2.5mm,size=title]
\textbf{Finding 3.}  \textit{ Functional stability is more brittle than representational structure under pruning.}
\label{question1}
\end{tcolorbox}

For all models, the LMC Barrier (top panel) begins to rise at much lower pruning levels and increases far more sharply than the CKA (bottom panel) drops. This is particularly noticeable in the Vision Transformers, where the LMC barrier soars to over $40\%$ accuracy drop at $40\%$ sparsity, while the CKA remains above $0.7$. This demonstrates that a model's functional equivalence to its unpruned self is rapidly lost under stress, even as the core geometry of its representations remains largely intact. Functional solutions, particularly for high-performance models, appear to be highly brittle and confined to narrow basins, whereas the underlying representational structure is a more robust, distributed property.

\subsection{Synthesizing the Triangle: Statistical Validation}
A core premise of our framework is that synthesizing multiple views yields deeper insights for scientific validation. We conduct a cross-view analysis correlating the static and sparsity views across 21 model pairs on the ImageNetV2 testbed. Our key findings are summarized in Table \ref{tab:stats_summary_new} and visualized in the Appendix (Figure \ref{fig:Cross_View}).

\begin{tcolorbox}[colback=blue!2!white,leftrule=2.5mm,size=title]
\textbf{Finding 4.} \textit{Static similarity and robustness under sparsity are strongly correlated, but deviations reveal critical insights.}
\end{tcolorbox}
Our analysis reveals a strong positive correlation between a model pair's static similarity and its robustness under sparsity (Pearson’s $r=0.882$, $p < 0.0001$).  We also found that static metrics can disagree; 4 of 21 pairs showed significant divergence between CKA and Procrustes, highlighting the risk of relying on a single metric. This discovery, summarized in Table \ref{tab:stats_summary_new}, shows that synthesizing views is essential for avoiding incorrect conclusions about model relationships. All other 20 pairs exhibited "Standard" patterns, where robustness was consistent with static similarity.

\begin{table}[h!]
\centering
\small
\caption{Statistical summary of Cross-View analysis. The table quantifies the correlation between views and the rate of metric disagreement.}
\label{tab:stats_summary_new}
\begin{tabular}{@{}lll@{}}
\toprule
\textbf{Analysis Category} & \textbf{Metric / Finding}  \\ \midrule
\multicolumn{3}{l}{\textit{Cross-View Correlation (static vs sparsity)}} \\
\quad Pearson Correlation & $r = 0.882$ \\ \midrule
\multicolumn{3}{l}{\textit{Metric Disagreement Analysis}} \\
\quad High disagreement cases & $4$ of $21$ ($19.0\%$) \\ \midrule
\end{tabular}
\end{table}

\subsection{The Triangle of Similarity in Practice}

Applying the full Triangle framework provides a holistic fingerprint for any model pair. While each pair generates a unique 1x3 panel figure, we describe the patterns observed for two illustrative cases.

\paragraph{Case 1: Same Architecture (e.g., ResNet18 vs. ResNet50).}
\begin{itemize}
    \item \textit{Panel 1 (Static):} Shows moderate-to-high CKA similarity with a clear diagonal, indicating good layer-to-layer correspondence.
    \item \textit{Panel 2 (Functional):} The LMC plot shows a small accuracy drop in the middle of the interpolation path, indicating the models are in nearby but distinct low-loss basins.
    \item \textit{Panel 3 (Sparsity):} Cross-model similarity often remains stable or slightly increases at moderate pruning levels before declining, suggesting pruning removes weights specific to each training run while preserving a shared architectural core.
\end{itemize}

\paragraph{Case 2: Different Architectures (e.g., ResNet18 vs. ViT-TP16).}
\begin{itemize}
    \item \textit{Panel 1 (Static):} Shows low baseline CKA, consistent with Figure \ref{fig:landscape}, with a less defined diagonal.
    \item \textit{Panel 2 (Functional):} LMC is not applicable. The Predictive Similarity score provides a single value indicating how closely their output distributions match, offering a functional comparison where LMC is impossible.
    \item \textit{Panel 3 (Sparsity):} The cross-model similarity curve under pruning often shows more complex, non-monotonic behavior. An initial drop may occur as architecture-specific features are removed, followed by a potential rise at extreme sparsity as both models are reduced to a minimal set of generic features.
\end{itemize}

\section{Discussion and Limitations}

While the Triangle of Similarity provides a robust framework for model comparison, several practical considerations remain.

\textbf{Computational Cost:} The framework is computationally demanding. CKA scales quadratically with samples ($O(N^2)$), and LMC requires multiple evaluations along high-dimensional paths, which can be expensive for large foundation models.

\textbf{Pruning Method:} We employ global magnitude pruning. Structured pruning, lottery ticket rewinding, or learned sparsity patterns may reveal different similarity dynamics and should be explored in future iterations of this framework.

\textbf{Practical Implications:} Despite these limitations, the multi-view approach guides decision-making. A model pair exhibiting high static similarity but low sparsity robustness (brittle core) indicates that while representations align at rest, one model may be significantly less transferable than the other.

\section{Conclusion and Future Work}

We propose and validate the Triangle of Similarity, a framework for comparing neural networks from static, functional, and dynamic perspectives, providing a holistic perspective on model similarity. Our initial application to a range of vision models highlights that architectural families form distinct representational clusters and that a model's task performance is often more brittle under pruning than its core representational structure. Our work provides a backbone for model comparison, demonstrating a strong correlation between static representational similarity and robustness under stress. 

\bibliography{references}

\appendix
\section{Statistical Analysis Visualization}
\label{sec:appendix}

\begin{figure}[h!]
\centering
\includegraphics[width=0.9\columnwidth]{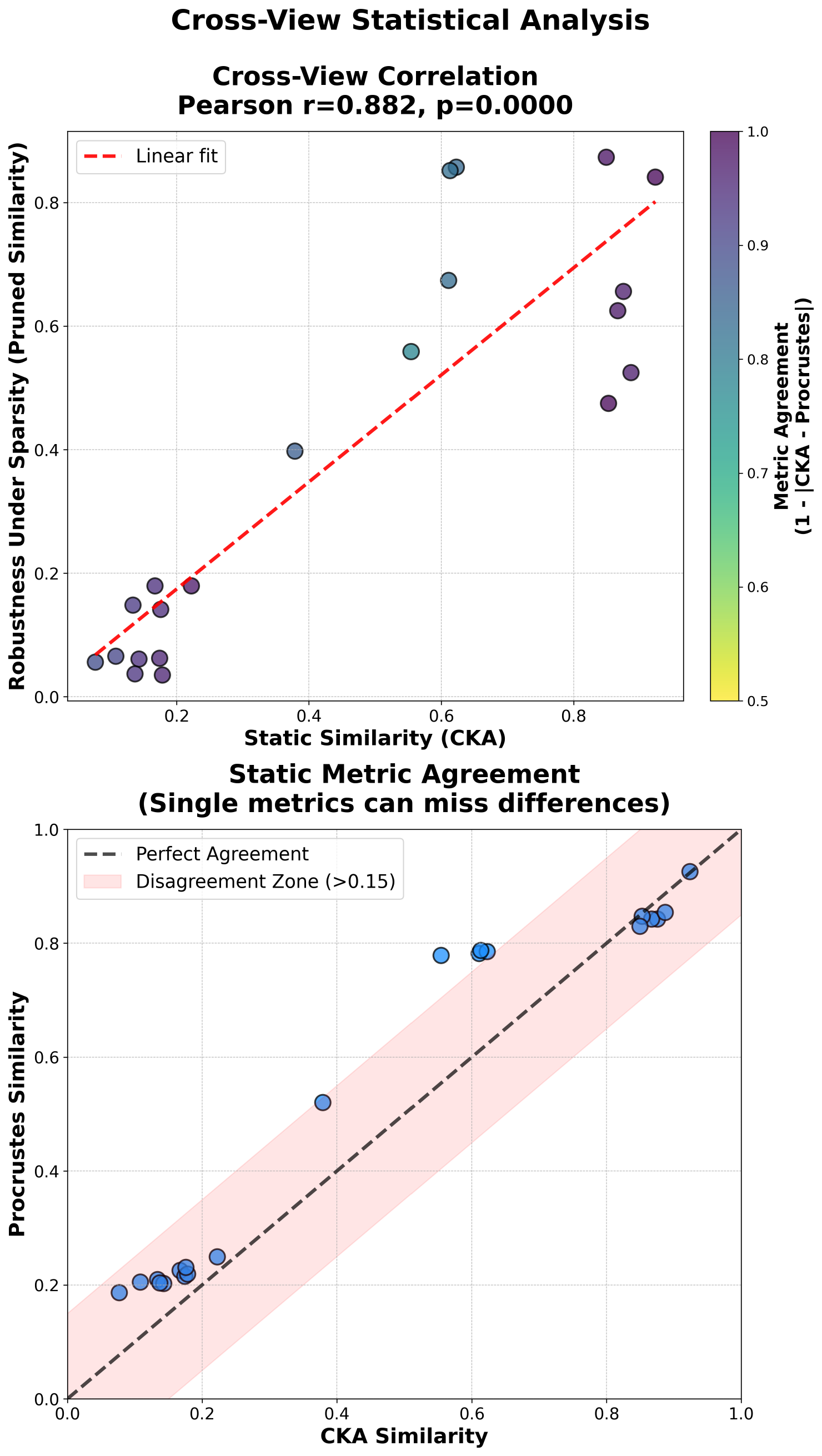} 
\caption{\textbf{Cross-View Statistical Analysis.} \textbf{Top:} A strong positive correlation (Pearson's $r=0.882$) exists between static similarity and robustness under sparsity. \textbf{Bottom:} CKA and Procrustes scores, while correlated, show several cases of high disagreement ($>0.15$), confirming that a multi-metric view is necessary.}
\label{fig:Cross_View}
\end{figure}

Figure \ref{fig:Cross_View} provides the detailed visual evidence for the statistical findings presented in the main text's "Synthesizing the Triangle" section.

\paragraph{Top Panel: Cross-View Correlation} This plot visualizes the strong positive correlation (Pearson's r=0.882) between the static similarity of a model pair (x-axis) and its robustness under sparsity (y-axis). The linear relationship confirms that, for the most part, models that start out more similar tend to have a more robust shared core.

\paragraph{Bottom Panel: Static Metric Agreement} This plot compares two different static metrics, CKA and Procrustes. While they are highly correlated, several data points fall into the shaded "Disagreement Zone," where the metrics differ by more than 0.15. This confirms that relying on a single static metric can be risky, justifying a multi-faceted approach even within a single view of the Triangle.

\end{document}